\documentclass[11pt,letterpaper]{llncs}
\usepackage{amsmath}
\usepackage{amssymb}
\usepackage{graphicx}
\usepackage{marvosym}
\usepackage{url}
\usepackage{fancyhdr}
\usepackage[T1]{fontenc}
\usepackage{comment}
\usepackage{subcaption}
\usepackage[sorting=none]{biblatex}

\usepackage{geometry}
\fancypagestyle{firstpage}{\fancyhf{}
\setcounter{page}{1}
\rfoot{\thepage}
}

\bibliography{biblio}

\begin{document}


\title{\LARGE{EXPLORING THE INFLUENCE OF RELEVANT KNOWLEDGE FOR NATURAL LANGUAGE GENERATION INTERPRETABILITY}}


%
%
\author{\large Iván Martínez-Murillo \and Paloma Moreda \and Elena Lloret}

\institute{\large University of Alicante \\%
Carr. de San Vicente del Raspeig, s/n \\%
San Vicente del Raspeig, Alicante, Spain, 03690 \\[0.5em]
\texttt{ivan.martinezmurillo@ua.es}}

%


%
%


\maketitle

\thispagestyle{firstpage}

\section*{ABSTRACT}
This paper explores the influence of external knowledge integration in Natural Language Generation (NLG), focusing on a commonsense generation task. We extend the CommonGen dataset by creating KITGI, a benchmark that pairs input concept sets with retrieved semantic relations from ConceptNet and includes manually annotated outputs. Using the T5-Large model, we compare sentence generation under two conditions: with full external knowledge and with filtered knowledge where highly relevant relations were deliberately removed. Our interpretability benchmark follows a three-stage method: (1) identifying and removing key knowledge, (2) regenerating sentences, and (3) manually assessing outputs for commonsense plausibility and concept coverage. Results show that sentences generated with full knowledge achieved 91\% correctness across both criteria, while filtering reduced performance drastically to 6\%. These findings demonstrate that relevant external knowledge is critical for maintaining both coherence and concept coverage in NLG. This work highlights the importance of designing interpretable, knowledge-enhanced NLG systems and calls for evaluation frameworks that capture the underlying reasoning beyond surface-level metrics.

\section*{KEYWORDS}
Natural Language Generation, Interpretability, Knowledge-enhanced, Commonsense Generation.


\section{INTRODUCTION}

Natural Language Generation (NLG) models have witnessed substantial advancements with the emergence of Transformer-based architectures \cite{vaswani2017attention}. The scaling of these models in terms of both size and training data has led to significant improvements in their performance in a wide range of downstream tasks \cite{yang2024giveus}. However, despite these advancements, recent research \cite{liu2023evaluating, hu2024towards, chang2024how} has highlighted persistent limitations in the ability of Large Language Models (LLMs) to store and generate factually accurate information \cite{miro2025roadmap}. These deficiencies pose significant challenges, particularly in domains where factual correctness is crucial, such as scientific writing, medical documentation, or legal reasoning.

To address this issue, a growing research direction focuses on integrating external knowledge sources into NLG models to enhance their factual consistency \cite{Jiang2024}. By leveraging structured knowledge bases, retrieval mechanisms, or hybrid neural-symbolic approaches, researchers aim to supplement the intrinsic knowledge of these models with verifiable external facts. This strategy has the potential to improve factual accuracy and contextual coherence in generated text. However, a critical challenge arises in evaluating the effectiveness of such knowledge integration techniques.

Current knowledge-enhanced NLG methods lack transparency analyses that explain how external knowledge contributes to performance improvements. Most existing approaches rely on automatic evaluation metrics, which mainly measure surface-level lexical similarity and often fail to accurately assess factual correctness in open-ended text generation tasks \cite{martinez2024analysing}.
Additionally, while some studies include manual evaluations, these typically focus only on the perceived quality of the generated texts and do not provide a deeper interpretability analysis of how external knowledge influences model behavior and output quality. This methodological gap highlights the need for a more comprehensive evaluation framework that extends beyond automated and manual metrics to incorporate interpretability analyses of the injected knowledge.

Therefore, this paper aims to address this gap by conducting a detailed interpretability analysis of how the quality of injected external knowledge influences NLG systems. The hypothesis is that enhancing NLG systems with non-related, or wrong external knowledge, critically affects their outputs. Specifically, we focus on a constrained commonsense generation task, enhanced with retrieved external knowledge, to evaluate commonsense reasoning in text generation. Our study investigates how knowledge integration affects the factual accuracy of generated text and examines the interpretability of these effects.

The contributions of this paper are twofold:
\begin{itemize}

\item To propose an extension to a widely-used commonsense reasoning dataset. We augment the dataset by incorporating: (1) external knowledge aligned with the input data, (2) automatically generated outputs conditioned on that knowledge, and (3) manually annotations of the generated sentences as either plausible or implausible. We named this resulting dataset as KITGI: Knowledge-Improved Text Generation and Interpretability. 

\item To propose a method and conduct a clear and detailed interpretability analysis of commonsense generation, demonstrating how the inclusion or removal of external knowledge influences the generated outputs.

\end{itemize}

By addressing these objectives, this work contributes to a more reliable assessment of knowledge-enhanced NLG models, offering insights into their factual generation capabilities and evaluation methodologies.  It complements existing automatic and human evaluation practices commonly employed in the field.

\section{RELATED WORK}

NLG field has advanced significantly with the introduction of the Transformer architecture \cite{vaswani2017attention}, greatly improving fluency and coherence. These models outperformed earlier approaches on complex language tasks, such as paraphrasing, question answering or machine translation \cite{riyadh2023towards}. As a result, NLG systems are now targeting more specific and demanding applications \cite{Jiang2024}. To support this, recent research focuses on integrating external knowledge to enhance factual accuracy and contextual relevance.

\textbf{Knowledge-Enhanced NLG}: It refers to the integration of external knowledge from diverse sources into NLG systems \cite{Jiang2024}. Techniques such as retrieval-augmented generation (RAG) \cite{rag2020}, knowledge-graph based generation \cite{liu-etal-2024-knowledge-graph}, or knowledge enhanced prompt tuning \cite{an-etal-2024-knowledge} have been shown to improve the comprehension and generative capabilities of NLG models.
Those methods enrich the models with additional context or facts retrieved from external sources such as knowledge graphs, domain-specific databases or documents. 

\textbf{NLG interpretability}: While knowledge-enhanced NLG improves the relevance and coherence of generated outputs, the mechanisms through which external information shapes the generation process remain insufficiently understood. Amnesic Probing \cite{amnesic2021} addresses this gap by using counterfactual examples to analyze the causal influence of injected knowledge on model predictions. ReX \cite{Liu_Zhang_2025} extend local, model-agnostic explanation techniques with temporal information, enhancing the alignment between input content and model outputs. Similarly, another approach \cite{feustel-etal-2024-enhancing} improves performance and interpretability by incorporating structured domain knowledge directly into dialogue systems, thereby increasing model transparency. Despite these advancements, more systematic methods are still needed to trace and quantify the influence of external knowledge on generation decisions, especially in contexts requiring commonsense reasoning. This research aims to address to mitigate this limitation.

\section{STARTING SETUP}
\label{sec:initialsetup}

To analyze the impact of integrating external knowledge into NLG systems, we propose and create the initial setup described below.

As we aim to analyze the effect of external knowledge on a constrained commonsense generation task, we modified and enriched a subset of the CommonGen dataset \cite{lin-etal-2020-commongen}, which is the most widely used dataset for this task. It involves generating a sentence that incorporates a given set of concepts to describe an everyday scenario. 

We began with 993 instances from the validation set of the CommonGen dataset and automatically generated sentences for each concept. These sentences were generated using T5-Large model \cite{raffelt5}. T5 is an encoder-decoder model pre-trained on a multi-task mixture of unsupervised and supervised tasks, and for which each task is converted into a text-to-text format. T5 works well on a variety of tasks out-of-the-box by prepending a different prefix to the input corresponding to each task. Furthermore, this model has shown remarkable performance on the CommonGen task\footnote{https://inklab.usc.edu/CommonGen/leaderboard.html}, as many of the approaches using this model as the foundational model obtained a great score with the automatic evaluation metrics.

The sentences were generated under two conditions:
\begin{itemize} 
\item No External Knowledge: The model generated a sentence using only the provided concept set, without any additional context.
\item Enhanced with External Knowledge: For each concept, we retrieved the top five semantic relations from ConceptNet \cite{speer2017conceptnet}, a knowledge graph where nodes represent concepts and edges represent relations such as ``is a part of'', ``used for'', or ``capable of''. These relations were appended to each instance alongside the original concept set, and the model generated a sentence using both the concepts and the supplementary knowledge. \end{itemize}

\begin{figure}
    \centering
    \includegraphics[width=1\linewidth]{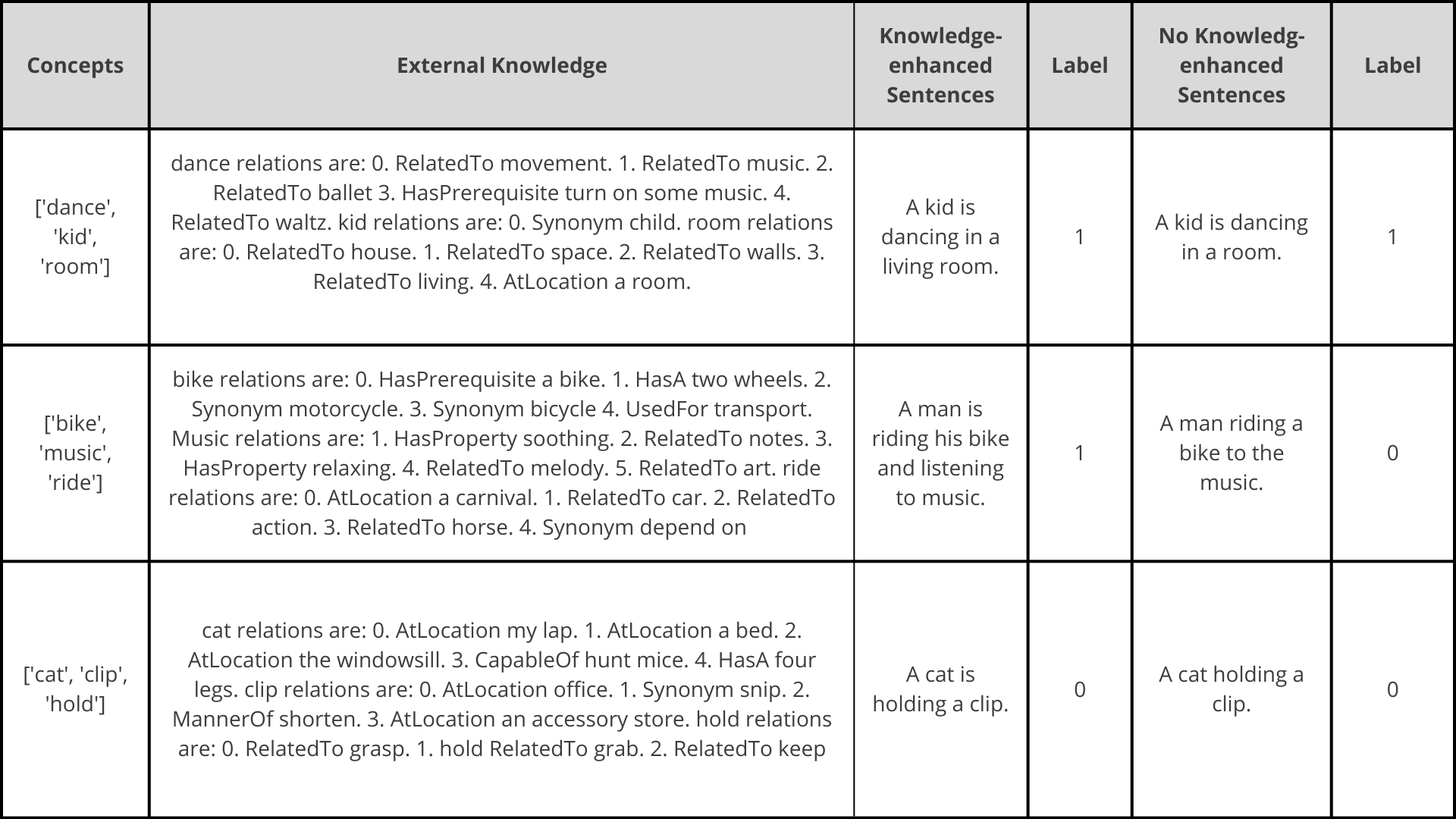}
    \caption{Samples from the crafted dataset.}
    \label{fig:commongen}
\end{figure}

Each generated sentence was then manually annotated as correct and plausible (1) or incorrect/implausible (0). Furthermore, for the sentences enhanced with external knowledge, the subset also incorporated the relations extracted from ConceptNet.
Figure \ref{fig:commongen} shows an example from the crafted dataset.

Notably, 121 sentences that were initially incorrect became correct after incorporating external knowledge. Therefore, our corpus for this study will be the 121 sentences that were improved after the injection of knowledge, alongside the corresponding concept sets and the retrieved knowledge from ConceptNet.

\section{INTERPRETABILITY METHOD}
The proposed interpretability benchmark consists of a three-stage method. First, key knowledge is removed; second, the model is retrained and the sentences are regenerated; and third, the results are manually labeled and evaluated. This process allows us to analyze the effects of different types of commonsense knowledge. The outcome is a final dataset of automatically generated sentences, labeled according to whether they contain commonsense or not. Each stage is explained in detail below.

\subsection{Stage 1: Analysis and Removal of Key Knowledge}
To assess the impact of external knowledge on the output of the NLG model, we focused on the subset of 121 sentences (i.e., evaluation dataset) for which the knowledge-enhanced model produced plausible outputs, which has been described in Section \ref{sec:initialsetup}.
Since the dataset includes the external knowledge used to enhance the model, we could directly determine its influence. To do so, we manually analyzed all the relations for each concept and discarded the ones that, according to human reasoning, could positively influence during the generation. With that, our goal was to corroborate that the injected knowledge really influenced the generation.  

\begin{figure}[htbp]
    \centering
    \begin{subfigure}[b]{0.45\linewidth}
        \centering
        \includegraphics[width=\linewidth]{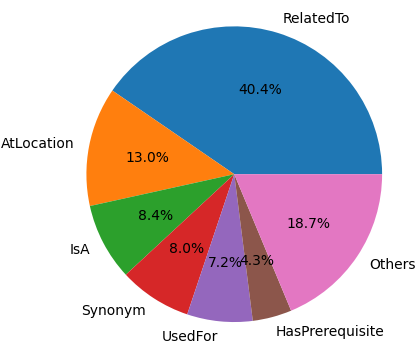}
        \caption{Distribution of relation types in the initial evaluation dataset.}
        \label{fig:relationsfull}
    \end{subfigure}
    \hfill
    \begin{subfigure}[b]{0.45\linewidth}
        \centering
        \includegraphics[width=\linewidth]{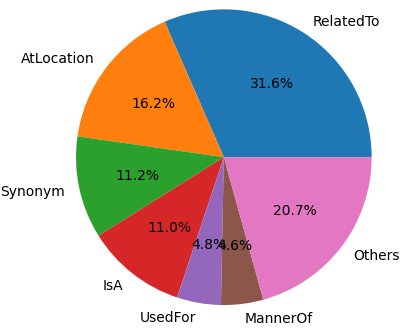}
        \caption{Distribution of relation types in the initial evaluation dataset after filtering out relevant relations.}
        \label{fig:relationspartial}
    \end{subfigure}
    \caption{Comparison of relation type distributions.}
    \label{fig:relationcomparison}
\end{figure}

We calculated the external knowledge included in the 121 evaluation sentences and found that in total they contained $1635$ relations corresponding to $121$ concept sets, each containing from $3$ to $5$ words. However, not every word yielded $5$ extractable relations.
With respect to the existing relations for those $121$ sentences of the dataset, the most common type found is the ``RelatedTo'' ($40.4\%$), followed by the ``AtLocation'' relations ($13.0\%$) and ``IsA'' ($8.4\%$). The total distribution of the relations can be seen in Figure \ref{fig:relationsfull}.

After calculating the total number of relations for the concepts, we carefully analyzed the relations related associated to the 121 concepts set. For each instance, we identified and removed the relations that, based on human judgment, seemed most useful for generating a sentence.

During this process, we removed $659$ relations ($40\%$ of the initial knowledge) that were deemed highly relevant to the given keywords based on human judgment. This process resulted in a final set of $976$ remaining relations. The distribution of relation types following this filtering step is shown in Figure \ref{fig:relationspartial}.

The most frequent relation type remained RelatedTo, which now accounts for $31.6\%$ of the dataset. However, this represents a $10\%$ decrease compared to its original proportion, suggesting that RelatedTo relations are particularly important from a human perspective when generating meaningful sentences.
The second most common relation type is AtLocation, comprising $16.2\%$ of the dataset. This is followed by the Synonym relation, which increased from $8.0\%$ to $11.2\%$ after filtering. This increase indicates that Synonym relations are less relevant to the specific contexts considered in our experiments and were therefore retained more frequently.

For example, consider the concept set \{\textit{``look'', ``watch'', and  ``window''}\}. The dataset includes the following relations: \{\textit{look relations are: 0. RelatedTo see. 1. RelatedTo glance. 2. RelatedTo	eyes. 3. RelatedTo seeing. 4. RelatedTo	view. watch relations are: 0. RelatedTo	time. 1. RelatedTo	wrist. 2. RelatedTo	clock. 3. RelatedTo look 4. RelatedTo clook. window relations are: 0. RelatedTo glass. 1. RelatedTo opening. 2. RelatedTo looking. 3. RelatedTo house. 4. RelatedTo wall.}\}.
We evaluated which of these relations would strongly influence sentence generation. For ``look'', we removed relations like see and seeing because they are highly relevant to the intended meaning and could make sentence generation easier.
In the case of ``watch'', most of its relations are associated with the concept of time, which is not relevant in this context. Therefore, we only removed the relation that connects ``watch'' with ``look''.
For ``window'', we found that the relation ``looking'' was most directly connected to the action we were focusing on, so we removed it as well. The remaining relations were kept to maintain background knowledge related to the object.

Once the relevant knowledge was filtered out, we generated the sentences again using the T5-Large model, incorporating the filtered knowledge as external input. Then, for this experiment, we analyze the following two sets:
\begin{itemize}
    \item A set of $121$ sentences generated using the complete set of relations as external knowledge. All of these sentences were manually labeled as correct, as stated in Section \ref{sec:initialsetup}.
    \item Another set of $121$ sentences generated using filtered knowledge, in which the most meaningful relations were intentionally excluded.
\end{itemize}

\subsection{Stage 2: Assessment of Commonsense and Coverage}
\label{subsec:evalmetrics}
We conducted a manual assessment based on two distinct criteria to evaluate the generated sentences after removing key knowledge. Each criterion was rated on a binary scale: a score of $0$ indicated inadequate performance, while a score of $1$ signified correct execution. The evaluation was carried out according to the following criteria:
\begin{itemize}
    \item \textbf{Commonsense:} We decide whether the generated sentence makes sense, or it does not make sense. 
    \item \textbf{Coverage:} We analyze if the generated sentence contains all the concepts in the concept set. 
\end{itemize}

\begin{figure}
    \centering
\includegraphics[width=\linewidth]{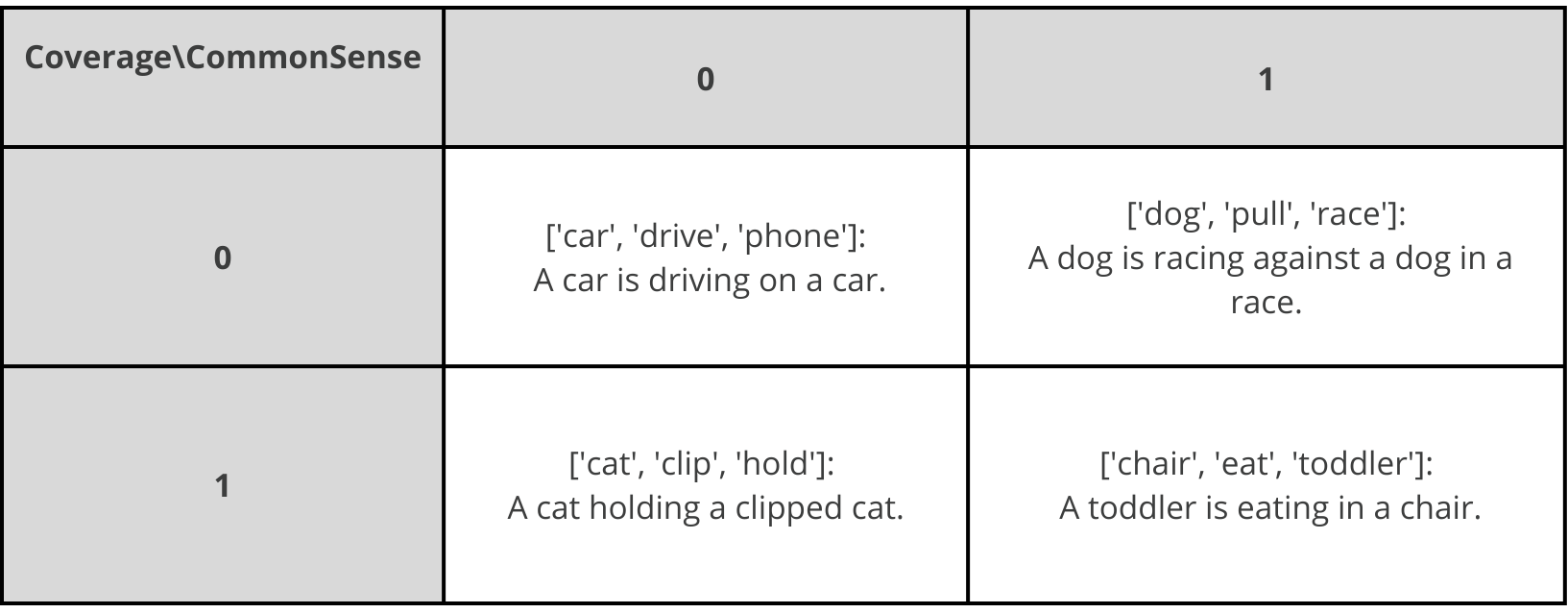}
    \caption{Representative samples of the criteria applied during evaluation.}
    \label{fig:evaluationsamples}
\end{figure}

Figure \ref{fig:evaluationsamples} presents an example of the evaluation process. The table is organized such that the vertical axis (rows) corresponds to the coverage criterion, while the horizontal axis (columns) reflects the commonsense criterion. We included the coverage dimension because, in many cases, the generated sentence appeared plausible yet failed to include all required keywords, thereby not fulfilling the task’s objective properly.

For instance, as it is shown in the figure, given the concepts ``dog'', ``pull'', and ``race'', the sentence ``A dog is racing against another dog in a race.'' is plausible and thus receives a commonsense score of $1$. However, since it omits the keyword ``pull'', it scores $0$ for coverage.

Some sentences fail on both criteria—lacking commonsense and omitting key terms. For example, for the concepts ``car'', ``drive'', and ``phone'', a sentence that excludes ``phone'' and is incoherent would receive a $0$ for both criteria.

Conversely, a sentence might include all keywords (coverage score of 1) but still lack coherence, resulting in a commonsense score of $0$. Thus, each criterion captures a distinct but complementary aspect of the sentence quality.

\subsection{Stage 3: KITGI Dataset Creation}
%
The final dataset proposed for this experiment comprises 121 instances, each containing the following components:

\begin{itemize} 
\item \textbf{Concept Set}: A group of 3 to 5 concepts. 
\item \textbf{Sentence with Full External Knowledge and Annotation}: A sentence generated by a T5-Large model, augmented with the complete set of retrieved knowledge, along with its annotation in terms of commonsense relevance and concept coverage. 
\item \textbf{Sentence with Filtered External Knowledge and Annotation}: A sentence generated by a T5-Large model, enhanced using only the filtered knowledge, with corresponding annotations for commonsense reasoning and concept coverage. 
\item \textbf{Retrieved Knowledge}: The set of relations retrieved from ConceptNet for each word in the concept set.
\item \textbf{Filtered Knowledge}: A subset of the retrieved knowledge, containing only the relations that are not relevant to each specific concept set. 
\end{itemize}

The dataset is available at \url{https://github.com/imm106/KITGI}.

\section{RESULTS AND DISCUSSION}
In this section, we present the scores from the manual evaluation conducted on the KITGI dataset, based on the criteria defined in Section \ref{subsec:evalmetrics}, and provide an analysis of the results.

\begin{figure}[htbp]
    \centering
    \begin{subfigure}[b]{0.45\linewidth}
        \centering
        \includegraphics[width=\linewidth]{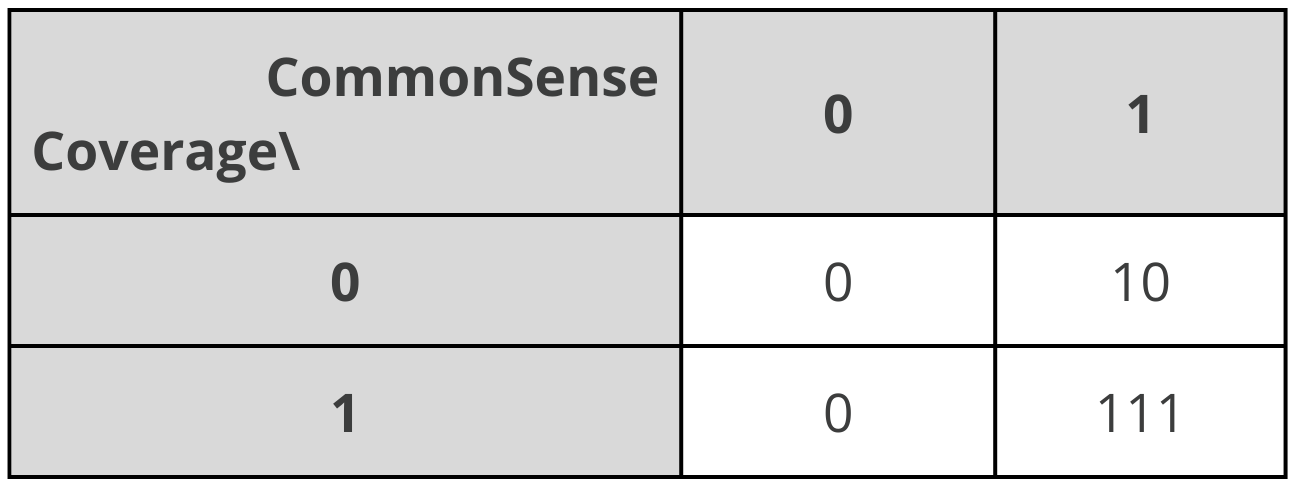}
        \caption{Manual analysis results for the sentences generated with all the knowledge.}
        \label{fig:manualfull}
    \end{subfigure}
    \hfill
    \begin{subfigure}[b]{0.45\linewidth}
        \centering
        \includegraphics[width=\linewidth]{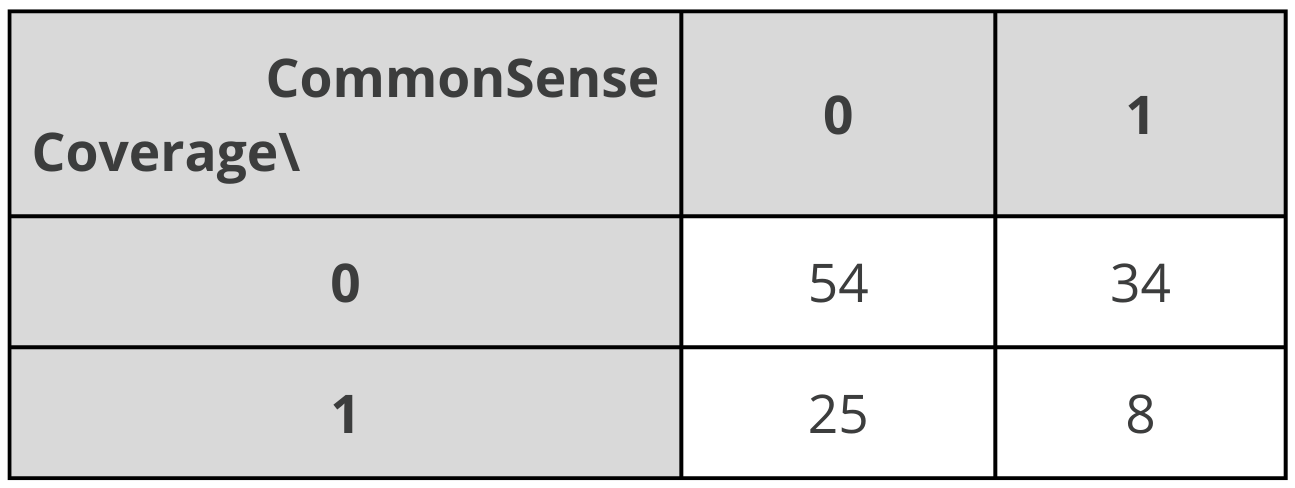}
        \caption{Manual analysis results for the sentences generated with all the filtered knowledge.}
        \label{fig:manualpartial}
    \end{subfigure}
    \caption{Manual analysis results.}
    \label{fig:manualeval}
\end{figure}

The outcomes of the manual evaluation are shown in Figure \ref{fig:manualeval}. Subfigure \ref{fig:manualfull} (left side) displays the results for sentences generated using the full external knowledge set, while Subfigure  \ref{fig:manualpartial} the right side shows the results for sentences generated after removing relevant knowledge (i.e., Filtered External Knowledge).

The initial set of sentences enhanced with the full external knowledge contained all commonsense, as described in Section \ref{sec:initialsetup}. However, as Subfigure \ref{fig:manualfull} shows, $8\%$ (10 sentences) of these did not include all the required keywords, meaning they did not fully meet the task's objective. In contrast, Subfigure \ref{fig:manualpartial} shows that only $42$ ($34 + 8$) sentences generated with filtered knowledge were considered meaningful - this corresponds to $34\%$ of the sentences. The performance is even lower when considering keyword coverage: only $8$ out of the $42$ meaningful sentences used all the words from the concept set. This represents just a $6\%$ of the dataset—a considerably low result compared to the $91\%$ of sentences that met both the coverage and commonsense criteria when enhanced with the full set of knowledge.

These results suggest that excluding relevant knowledge significantly impacts the coverage criterion. Specifically, in Subfigure \ref{fig:manualpartial} is shown that $88$ ($54 + 34$) of the $121$ sentences ($72\%$) did not include at least one required concept word. Of the remaining $33$ sentences that included all concept words, only $8$ were complete and meaningful, thus having commonsense.

Indeed, after carefully analyzing the generated sentences with the filtered knowledge, we detected three variants: 
\begin{itemize}
    \item \textbf{The external knowledge associated with certain words is misleading}: When this happens, the system often omits the problematic word in the generated sentence, sometimes still producing a sentence that is meaningful. This suggests that the model struggles to integrate the word with the context provided by the external knowledge and chooses to exclude it instead. For example, consider the concept set [``look'', ``watch'', ``window'']. The external knowledge provided for the word ``watch'' refers only to the object (e.g., a wristwatch), rather than its verb form. As a result, the model is unable to combine ``watch'' meaningfully with the other words, and generates the sentence ``A man is looking at a window.'' Another example involves the concept set [``fall'', ``ground'', ``jump''], where the model generates “A man is jumping on the ground.” In this case, the knowledge for the word “fall” is related to the season (Autumn), rather than the verb ``to fall,'' preventing the model from using it correctly in the intended context.
    \item \textbf{The external knowledge is not helpful}: In many cases, although the provided knowledge corresponds to the correct meaning of each individual word in context, it does not support establishing meaningful connections between the words in the concept set. As a result, the system often generates nonsensical sentences. In some cases, it also fails to include all the words from the concept set. An example of this can be seen with the concept set [``attempt'', ``fence'', ``knife'', ``stick'', ``throw'']. The model generates the sentence ``Someone throws a knife and attempts to throw it into the fence.'' While the external knowledge is relevant for each word—``attempt'' is associated with trying, ``fence'' refers to a protective wall, ``knife'' is defined as a cutting object, ``stick'' as a small piece of wood, and ``throw'' as the action of launching something—there are no strong semantic relations linking these concepts together. As a result, the sentence lacks coherence despite the relevance of the individual word meanings.
    \item \textbf{The given knowledge establishes a slight connection among words}: In some cases, the provided knowledge does not establish a direct relationship among the concepts, but it still indirectly helps the model generate a coherent and accurate sentence. For example, consider the concepts [``boat'', ``sail'', ``day'']. The knowledge includes that a ``boat'' can travel on water and is located on a lake; ``sail'' is associated with wind and cloth; and ``day'' is defined as the antonym of night or related to time. Based on this, the system generates the sentence: ``Boats sail on a sunny day.'' Here, the knowledge about boats supports the idea of traveling on water, which could help to connect it with the concept of sailing. Similarly, the knowledge about day helps form the phrase sunny day, drawing on its contrast with night. Thus, although the relationships are not explicitly defined, background knowledge still helps guide the model toward a meaningful output.
\end{itemize}

These findings demonstrate that the quality of external knowledge significantly impacts the model's performance. When the input includes misleading or ambiguous information, the model struggles to integrate it effectively, often resulting in incorrect sentence generation. This highlights the importance of considering the input context when retrieving relevant knowledge. Given the richness of language, many words have multiple meanings, and selecting the wrong one can negatively affect the output. Conversely, even a weak but relevant connection in the external knowledge can help the model produce a more accurate result.

\section{CONCLUSIONS \& FUTURE WORK}

This study presents an in-depth interpretability method and analysis of how external knowledge enhances NLG, specifically in a constrained commonsense reasoning task. Using a controlled benchmark, we systematically eliminated highly relevant semantic relations to assess their impact. The results reveal that properly integrated external knowledge is essential for producing coherent and plausible sentences. We found that removing critical knowledge elements markedly reduces both the commonsense accuracy and conceptual coverage of the generated outputs, highlighting the vital role of external knowledge in ensuring factually grounded and logically consistent language generation.

For future work, this research can be extended in several directions. First, we plan to investigate the impact of external knowledge in multilingual settings to determine whether its influence is consistent across languages or language-dependent. Second, this approach could be applied to other NLP tasks to assess whether knowledge integration yields similar effects beyond text generation. Finally, it would be valuable to explore more advanced knowledge retrieval and integration methods to evaluate how different types and sources of knowledge influence model performance.

\section*{ACKNOWLEDGEMENTS}
This research is part of the R\&D projects ``CORTEX: Conscious Text Generation'' (PID2021-123956OB-I00), funded by MCIN/AEI/10.13039/501100011033/ and by ``ERDF A way of making Europe''; QUMLAUDE: Mecánica cuántica para comprensión y generación del lenguaje'' (PID2024-160791OB-I00), funded by MCIN/AEI/10.13039/501100011033/ and by ``ERDF A way of making Europe; ``CLEAR.TEXT: Enhancing the modernization public sector organizations by deploying Natural Language Processing to make their digital content CLEARER to those with cognitive disabilities'' (TED2021-130707B-I00), funded by MCIN/AEI/10.13039/501100011033 and ``European Union NextGenerationEU/PRTR''; and project ``NL4DISMIS: Natural Language Technologies for dealing with dis- and misinformation with grant reference (CIPROM/2021/021)" funded by the Generalitat Valenciana.  
Additionally, this work is funded by the Ministerio para la Transformación Digital y de la Función Pública and Plan de Recuperación, Transformación y Resiliencia - Funded by EU – NextGenerationEU within the framework of the project Desarrollo Modelos ALIA.
Moreover, it has been also partially funded by the Ministry of Economic Affairs and Digital Transformation and ``European Union NextGenerationEU/PRTR'' through the ``ILENIA'' project (grant number 2022/TL22/00215337) and ``VIVES'' subproject (grant number 2022/TL22/00215334). 

\printbibliography

\vspace{2cm}

\section*{AUTHORS}
\noindent {\bf Iván Martínez-Murillo} is a Ph.D. student in Computational Linguistics at the University of Alicante (Spain). His primary research interests lie in natural language generation, controllable generation methods and hallucination mitigation techniques. Martínez-Murillo received a Master's degree in Cybersecurity from the University of Alicante in 2022. In 2023, he joined the Language Processing and Information Systems Group from the University of Alicante. \\Contact him at ivan.martinezmurillo@ua.es.\\

\noindent {\bf Paloma Moreda}  is a Lecturer in the Department of Languages and Computing Systems of the University of Alicante (Spain).  His research interests include Text Simplification, Gender Bias and , Automatic Language Generation and some other Natural Language Processing topics. She received the Ph.D. degree in Natural Language Processing, and has been the PI of several research project. Contact her at moreda@dlsi.ua.es.\\

\noindent {\bf Elena Lloret} is a Full Professor in the Department of Software and Computing Systems of the University of Alicante (Spain), teaching courses related to Databases and Natural Language Processing. Her research interests are in the field of Natural Language Processing, with special emphasis in Automatic Summarization and Natural Language Generation. Lloret received her Ph.D. degree in computer science from the University of Alicante in 2011. Contact her at elloret@dlsi.ua.es.

\end{document}